%% file: main.tex
\documentclass[10pt,twocolumn,letterpaper]{article}

\usepackage{cvpr}
\usepackage{times}
\usepackage{epsfig}
\usepackage{amsmath}
\usepackage{amssymb}

\usepackage[pagebackref=true,breaklinks=true,letterpaper=true,colorlinks,bookmarks=false]{hyperref}

\cvprfinalcopy

\input{math_commands.tex}

\usepackage{mathtools}  
\usepackage{multicol}
\usepackage{graphicx}
\usepackage{booktabs}
\usepackage{url}

\title{Domain Adaptive Transfer Learning with Specialist Models}

\author{
    Jiquan Ngiam, Daiyi Peng, Vijay Vasudevan, Simon Kornblith, 
    Quoc V. Le, Ruoming Pang\\
    Google Brain \\
    \texttt{\{jngiam,daiyip,vrv,skornblith,qvl,rpang\}@google.com} \\
}

\newcommand{\largedatasetname}{JFT}
\newcommand{\largedatasetcitep}{ \cite{Sun2017} }
\newcommand{\largedatasetcitet}{ \cite{Sun2017} }
\newcommand{\largedatasetfootnote}{}

\begin{document}
\maketitle

\begin{abstract}

Transfer learning is a widely used method to build high performing computer vision models. In this paper, we study the efficacy of transfer learning by examining how the choice of data impacts performance. We find that more pre-training data does not always help, and transfer performance depends on a judicious choice of pre-training data. These findings are important given the continued increase in dataset sizes. We further propose domain adaptive transfer learning, a simple and effective pre-training method using importance weights computed based on the target dataset. Our method to compute importance weights follow from ideas in domain adaptation, and we show a novel application to transfer learning. Our methods achieve state-of-the-art results on multiple fine-grained classification datasets and are well-suited for use in practice.

\end{abstract}

\section{Introduction}

Transfer learning using pre-trained models is one of the most successfully applied methods in the field of computer vision. In practice, a model is first trained on a large labeled dataset such as ImageNet \cite{Russakovsky2015}, and then fine-tuned on a target dataset. During fine-tuning, a new classification layer is learned from scratch, but the parameters for the rest of the network layers are initialized from the ImageNet pre-trained model. This method to initialize training of image models has proven to be highly successful and is now a central component of object recognition \cite{Razavian2014}, detection \cite{Girshick2015,Ren2015,Huang2017}, and segmentation \cite{Shelhamer2017,Chen2018,He2017}. 

By initializing networks with ImageNet pre-trained parameters, models converge faster, requiring less training time, and often achieve higher test accuracy. They can also achieve good performance when the target dataset is small.  Most prior work have considered only ImageNet as the source of pre-training data due its large size and availability. In this work, we explore how the choice of pre-training data can impact the accuracy of the model when fine-tuned on a new dataset.

To motivate the problem, consider a target task where the goal is to classify images of different food items (e.g., `hot dog' v.s. `hamburger') for a mobile application~\cite{Anglade2017}.  A straight-forward approach to applying transfer learning would be to employ an ImageNet pre-trained model fine-tuned on a food-specific dataset. However, we might wonder whether the pre-trained model, having learned to discriminate between irrelevant categories (e.g., `dogs' vs. `cats'), would be helpful in this case of food classification. More generally, if we have access to a large database of images, we might ask: is it more effective to pre-train a classifier on all the images, or just a subset that reflect food-like items? 

Furthermore, instead of making a hard decision when selecting pre-training images, we can consider a soft decision that weights each example based on their relevancy to the target task. This could be estimated by comparing the distributions of the source pre-training data and the target dataset. This approach has parallels to the covariate shift problem often encountered in survey and experimental design~\cite{Shimodaira2000}. 

We study different choices of source pre-training data and show that a judicious choice can lead to better performance on all target datasets we studied. Furthermore, we propose domain adaptive transfer learning - a simple and effective pre-training method based on importance weights computed based on the target dataset.

\subsection{Summary of findings}

\paragraph{More pre-training data does not always help.} We find that using the largest pre-training dataset does not always result in the best performance. By comparing results of transfer learning on different subsets of pre-training data, we find that the best results are obtained when irrelevant examples are discounted. This effect is particularly pronounced with fine-grained classification datasets. 

\paragraph{Matching to the target dataset distribution improves transfer learning.} We demonstrate a simple and computationally-efficient method to determine relevant examples for pre-training. Our method computes importance weights for examples on a pre-training dataset and is competitive with hand-curated pre-training datasets. Using this method, we obtain state-of-the-art results on the fine-grained classification datasets we studied (e.g., Birdsnap, Oxford Pets, Food-101).

\paragraph{Fine-grained target tasks require fine-grained pre-training.} We find that transfer learning performance is dependent on whether the pre-training data captures similar discriminative factors of variations to the target data. When features are learned on coarse grained classes, we do not observe significant benefits transferred to fine-grained datasets. 

\section{Related Work}

The success of applying convolution neural networks to the ImageNet classification problem \cite{Krizhevsky2012} led to the finding that the features learned by a convolutional neural network perform well on a variety of image classification problems \cite{Razavian2014,Donahue2014}. Further fine-tuning of the entire model was found to  improve performance \cite{Agrawal2014}.

Yosinski \etal~\cite{Yosinski2014} conducted a study of how transferable ImageNet features are, finding that the higher layers of the network tend to specialize to the original task, and that the neurons in different layers in a network were highly co-adapted. They also showed that distance between tasks matters for transfer learning and examined two different subsets (man-made v.s. natural objects). Azizpour \etal~\cite{Azizpour2016} also examined different factors of model design such as depth, width, data diversity and density. They compared data similarity to ImageNet based on the task type: whether it was classification, attribute detection, fine-grained classification, compositional, or instance retrieval. 

Pre-training on weakly labeled or noisy data was also found to be effective for transfer learning. Krause \etal~\cite{Krause2016} obtained additional noisy training examples by searching the web with the class labels. We note that our method does not use the class labels to collect additional data. Mahajan \etal~\cite{Mahajan2018} were able to attain impressive ImageNet performance by pre-training on 3 billion images from Instagram. Notably, they found that it was important to select appropriate hash-tags (used as weak labels) for source pre-training, and suggested a heuristic for their specific dataset. 

Understanding the similarity between datasets based on their content was studied by Cui \etal~\cite{Cui2018}, who suggest using the Earth Mover's Distance (EMD) as a distance measure between datasets. They constructed two pre-training datasets by selecting subsets of ImageNet and iNaturalist, and showed that selecting an appropriate pre-training subset was important for good performance. Ge and Yu~\cite{Ge17} used features from filter bank responses to select nearest neighbor source training examples and demonstrated better performance compared to using the entire source dataset. Zamir \etal~\cite{Zamir2018} define a method to compute transferability \textit{between tasks} on the same input; our work focuses on computing relationships \textit{between different input datasets}.

In a comprehensive comparison, Kornblith \etal~\cite{Kornblith2018} studied fine-tuning a variety of models on multiple datasets, and showed that performance on ImageNet correlated well with fine-tuning performance. Notably, they found that transfer learning with ImageNet was ineffective for small, fine-grained datasets. 

Our approach is related to domain adaptation which assumes that the \textit{training} and \textit{test} set have differing distributions~\cite{Shimodaira2000}. We adopt similar ideas of importance weighting examples~\cite{Sugiyama2007,Saerens2002,Zhang2013} and adapt them to the \textit{pre-training} step instead, showing that this is an effective approach. While our derivation of importance weights is similar to past work in domain adaptation, the application to the transfer learning setting is novel.

In this work, we show that transfer learning to fine-grained datasets is sensitive to the choice of pre-training data, and demonstrate how to select pre-training data to significantly improve transfer learning performance. We build on the work of \cite{Cui2018,Ge17}, demonstrating the effectiveness of constructing pre-training datasets. Furthermore, we present a simple, scalable, and computationally-efficient way to construct pre-training datasets.

\section{Transfer learning setup}

\begin{table}[!t]
\centering
\begin{tabular}{l c c} 
 \toprule
 Top Ancestor Label & \# Examples & \# Classes \\ 
 \midrule
 \textit{Entire Dataset} & 300M & 18,291 \\ 
 \hline \\[-0.7em]
 Animal & 33.5M & 2,992 \\ 
 Bird & 5.4M & 403  \\
 \hline \\[-0.7em]
 Car & 27.9M & 2,959 \\
 Aircraft & 3.1M & 418 \\ 
 Vehicle & 43.7M & 3,969 \\
 Transport & 45.0M & 3,987 \\
 \hline \\[-0.7em]
 Food & 18.4M & 3,532 \\
 \bottomrule
\end{tabular}
\vspace{1.0em}
\caption{\largedatasetname~\largedatasetcitep dataset statistics. We hand-picked several subsets by hand-selecting labels based on the target dataset. We compare our methods against these hand-picked subsets.}
\label{table:jft_subset_table}
\end{table}

We use the \largedatasetname\largedatasetfootnote\largedatasetcitep and ImageNet \cite{Russakovsky2015} datasets as our source pre-training data and consider a range of target datasets for fine-tuning (Section \ref{section:target_datasets}). For each target dataset, we consider different strategies for selecting pre-training data, and compare the fine-tuned accuracy. We do not perform any label alignment between the source and target datasets. During fine-tuning, the classification layer in the network is trained from random initialization. The following sections describe the datasets and experiments in further detail.

\begin{figure*}[!tbh]
  \centering
  \includegraphics[width=0.95\textwidth]{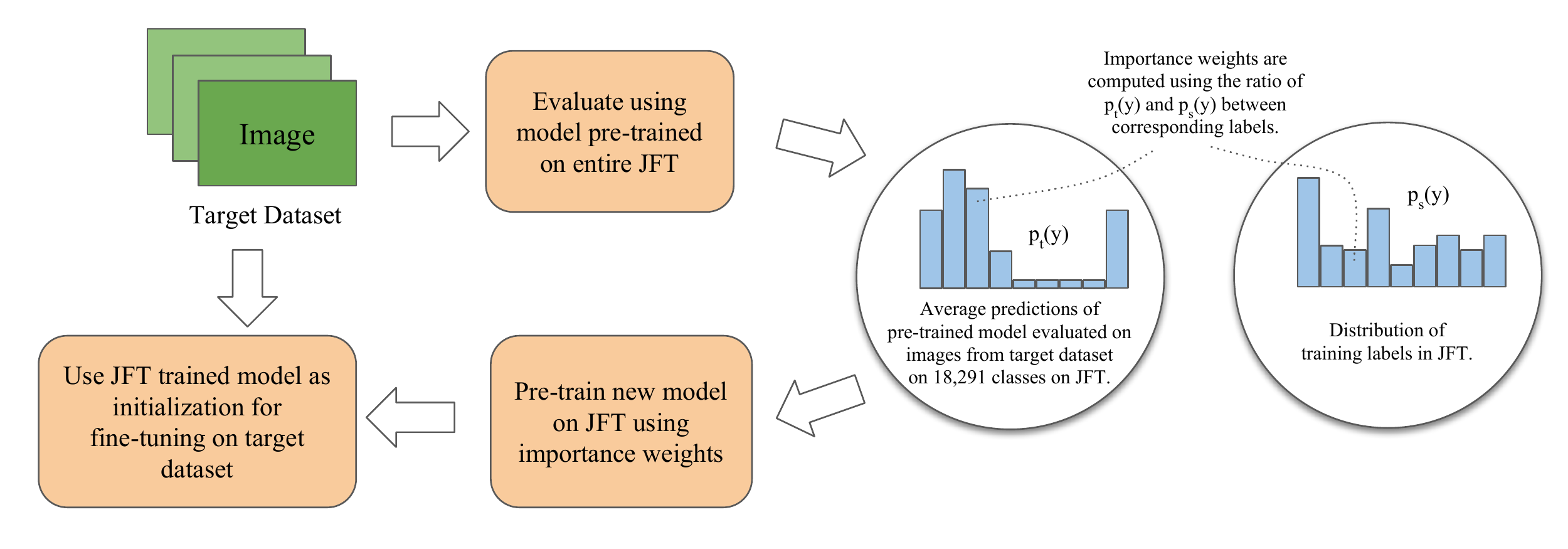}
  \caption{To compute importance weights, we start by evaluating images from the target dataset using an image model pre-trained on the entire JFT dataset. For each image, this gives us a prediction over the 18,291 classes in JFT. We average these predictions to obtain $p_t(y)$. We estimate $p_s(y)$ from the source pre-training dataset directly by dividing the number of times a label appears in the source pre-training dataset by the total number of examples in the source pre-training dataset. The ratio $p_t(y)/p_s(y)$ provides the importance weight for a given label in the source pre-training dataset. A model is pre-trained on the entire JFT dataset using these importance weights and then fine-tuned on the target dataset. }
  \label{fig:atl_diagram}
\end{figure*}

\subsection{Source pre-training data}

The \largedatasetname~dataset has 300 million images and 18,291 classes. Each image can have multiple labels and on average, each image has 1.26 labels. The large number of labels include many fine-grained categories, for example, there are 1,165 different categories for animals. While the labels are noisy and sometimes missing, we do not find this to a be a problem for transfer learning in practice. The labels form a semantic hierarchy: for example, the label `mode of transport' includes the label `vehicle', which in turn includes `car'. 

The semantic hierarchy of the labels suggests a straight-forward approach to constructing different subsets of \largedatasetname~as source pre-training data. Given a label, we can select all of its child labels in the hierarchy to form a label set, with the corresponding set of training examples. We created 7 subsets of \largedatasetname~across a range of labels\footnote{The following parent-child relationships exists in the label hierarchy: bird $\subset$ animal; car $\subset$ vehicle $\subset$ transport; aircraft $\subset$ vehicle $\subset$ transport. We note that \largedatasetcitet excluded classes with too few training examples during training, while we include all classes available.}  (Table \ref{table:jft_subset_table}).

However, creating subsets using the label hierarchy can be limiting for several reasons: (a) the number of examples per label are pre-defined by the \largedatasetname~dataset; (b) not all child labels may be relevant; (c) a union over different sub-trees of the hierarchy may be desired; and (d) not all source datasets have richly-defined label hierarchies.  In section \ref{section:importance_weighting}, we discuss a domain adaptive transfer learning approach that automatically selects and weights the relevant pre-training data.
\begin{table*}[!t]
\centering
\begin{tabular}{l c c c} 
 \toprule
 Target Dataset & \# Training Examples & \# Test Examples & \# Classes \\
 \midrule
 CIFAR-10 \cite{Krizhevsky2009} & 50,000 & 10,000 & 10 \\ 
 Birdsnap \cite{Berg2014} & 47,386 & 2,443 & 500  \\
 Stanford Cars \cite{Krause2013} & 8,144 & 8,041 & 196 \\
 FGVC Aircraft \cite{Maji2013} &  6,667 & 3,333 & 100 \\
 Oxford-IIIT Pets \cite{Parkhi2012} & 3,680 & 3,369 & 37 \\ 
 Food-101 \cite{Bossard2014} & 75,750 & 25,250 & 101 \\
 \bottomrule
\end{tabular}
\vspace{1.0em}
\caption{Target datasets for fine-tuning.}
\label{table:target_dataset}
\end{table*}

\subsection{Target training dataset}
\label{section:target_datasets}

We evaluate the performance of transfer learning on a range of classification datasets (Table \ref{table:target_dataset}) that include both general and fine-grained classification problems. Using the same method as Krause \etal~\cite{Krause2016}, we ensured that the source pre-training data did not contain any of the target training data by removing all near-duplicates of the target training and test data from the \largedatasetname~dataset\footnote{We used a CNN-based duplicate detector and chose a conservative threshold for computing near-duplicates to err on the side of ensuring that duplicates were removed. We removed a total of 48k examples from \largedatasetname, corresponding to duplicates that were found in target datasets.}.

\subsection{Domain adaptive transfer learning by importance weighting}
\label{section:importance_weighting}

In this section, we propose domain adaptive transfer learning, a simple and effective way to weight examples during pre-training. Let us start by considering a simplified setting where our source and target datasets are over the same set of values in pixels $x$, and labels $y$; we will relax this assumption later in this section. 

During pre-training, we usually optimize parameters $\theta$ to minimize a loss function $\mathbb{E}_{x, y \sim D_{s}} [ L(f_\theta(x), y) ]$ computed empirically over a source dataset $D_{s}$. $L(f_\theta(x), y)$ is often the cross entropy loss between the predictions of the model $f_\theta(x)$ and the ground-truth labels $y$. However, the distribution of source pre-training dataset $D_{s}$ may differ from the target dataset $D_{t}$. This could be detrimental as the model may emphasize features which are not relevant to the target dataset. We will mitigate this by up-weighting the examples that are most relevant to the target dataset. This is closely related\footnote{Prior work on prior probability shift usually considered shifts between train and test set, while we instead consider differences between the pre-training and training datasets.} to prior probability shift~\cite{Saerens2002,Storkey2009} also known as target shift~\cite{Zhang2013}.

We start by considering optimizing the loss function over the target dataset, $D_t$ instead: 

\begin{flalign*}  
\mathbb{E}_{x, y \sim D_{t}} \big[ L(f_\theta(x), y) \big]
& = \sum_{x,y} P_{t}(x, y) L(f_\theta(x), y) 
\end{flalign*}

where we use $P_s$ and $P_t$ to denote distributions over the source and target datasets respectively. We first reformulate the loss to include the source dataset $D_s$:

\begin{flalign*}
= \sum_{x,y} P_{s}(x, y) \frac{P_{t}(x, y)}{P_{s}(x, y)} L(f_\theta(x), y) 
\\
= \sum_{x,y} P_{s}(x, y) \frac{P_{t}(y)P_{t}(x|y)}{P_{s}(y)P_{s}(x|y)} L(f_\theta(x), y)
\end{flalign*}

Next, we make the assumption that $P_s(x|y) \approx P_t(x|y)$, that is the distribution of examples given a particular label in the source dataset is approximately the same as that of the target dataset. We find this assumption reasonable in practice: for example, the distribution of `bulldog' images from a large natural image dataset can be expected to be similar to that of a smaller animal-only dataset. This assumption also allows us to avoid having to directly model the data distribution $P(x)$. 

Cancelling out the terms, we obtain:

\begin{flalign*}
\label{eq:appoximation}
\approx \sum_{x,y} P_{s}(x, y) \frac{P_{t}(y)}{P_{s}(y)} L(f_\theta(x), y)  \\
= \mathbb{E}_{x, y \sim D_{s}} \big[ \frac{P_{t}(y)}{P_{s}(y)} L(f_\theta(x), y) \big]
\end{flalign*}

Intuitively, $P_t(y)$ describes the distribution of labels in the target dataset, and $P_t(y)/P_s(y)$ reweights classes during source pre-training so that the class distribution statistics match $P_t(y)$. We refer to $P_t(y)/P_s(y)$ as importance weights and call this approach of pre-training \textit{Domain Adaptive Transfer Learning}.

For this approach to be applicable in practice, we need to relax the earlier assumption that the source and target datasets share the same label space. Our goal is to estimate $P_t(y)/P_s(y)$ for each label in the source dataset. The challenge is that the source and target datasets have different sets of labels. Our solution is to estimate both $P_t(y)$ and $P_s(y)$ for labels in the source domain. The denominator $P_s(y)$ is obtained by dividing the number of times a label appears by the total number of source dataset examples. To estimate $P_t(y)$, we use a classifier to compute the probabilities of \textit{labels from source dataset} on \textit{examples from the target dataset}.

Our method is described in Figure \ref{fig:atl_diagram}. Concretely, we first train an image classification model on the entire source dataset. Next, we feed only the images from the target dataset into this model to obtain a prediction for each target example. The predictions are averaged across target examples, providing an estimate of $P_t(y)$, where $y$ is specified over the source label space. 
We emphasize that this method does not use the target labels when computing importance weights.

Our approach is in contrast to Ge and Yu~\cite{Ge17}, which is computationally expensive as they compute a similarity metric between every pair of images in the source dataset and target dataset. It is also more adaptive than Cui \etal~\cite{Cui2018}, which suggests selecting appropriate labels to pretrain on, without specifying a weight on each label. 

\section{Experiments}

\begin{table*}[!thb]
\centering
\begin{tabular}{l | c c c c c c} 
 \toprule 
 & \multicolumn{6}{c}{Target Dataset} \\
 Pre-training Method & Birdsnap & Oxford-IIIT & Stanford  & FGVC      & Food-101 & CIFAR-10 \\ 
         &          & Pets        & Cars      & Aircraft         &          & \\ 
 \midrule
 Entire \largedatasetname~Dataset &  74.2   &  92.5    & 94.0   &  88.2        &  88.6       &  97.6 \\
 \largedatasetname~- Bird       &  80.7 &  86.4    &  88.1   &  74.9      &  87.5       &  96.9   \\ 
 \largedatasetname~- Animal     &  77.8  & 96.7 &  89.1  &  78.2   &  89.2  &  98.1   \\ 
 \largedatasetname~- Car        &  73.4    &  79.8 &  \textbf{96.0}   &  82.1       &  86.1       &  93.0   \\ 
 \largedatasetname~- Aircraft   &  73.4    &  78.7 &  88.2 &  91.1 &  87.1       &  96.1   \\ 
 \largedatasetname~- Vehicle    &  74.2    &  79.6 &  95.8   &  86.8        &  81.6       &  96.4   \\ 
 \largedatasetname~- Transport  &  74.4    &  78.4 &  95.9   &  88.4        &  86.9       &  96.2   \\ 
 \largedatasetname~- Food       &  74.9    &  81.1 &  90.3   &  85.6        &  93.5 &  96.4   \\ 
 \largedatasetname~- Adaptive Transfer    &  \textbf{81.7}   &  \textbf{97.1}   &  95.7   &  \textbf{94.1}   &  \textbf{94.1}   &  \textbf{98.3}  \\
  \hline \\ [-0.8em]
 ImageNet - Entire Dataset &  77.2 &  93.3 &  91.5   &  88.8       &  88.7       &  97.4   \\ ImageNet - Adaptive Transfer &  76.6 &  94.1 &  92.1   &  87.8    &  88.9       &  97.7   \\ \hline \\ [-0.8em]
 Random Initialization &  75.2 &  80.8 &  92.1   &  88.3        &  86.4       &  95.7  \\
\bottomrule
\end{tabular} 
\vspace{1.0em}
\caption{Transfer learning results with Inception v3. Each row corresponds to a pre-training method. Adaptive transfer refers to our proposed method described in section \ref{section:importance_weighting}. Each column corresponds to one target dataset. Results reported are top-1 accuracy for all datasets except Oxford-IIIT Pets, where we report mean accuracy per class. All our results are averaged over 5 fine-tuning runs. Adaptive transfer is better or competitive with the hand selected subsets.\medskip}
\label{table:transfer_results}
\end{table*}

We used the Inception v3 \cite{Szegedy2016}, and AmoebaNet-B \cite{Real2018} models in our experiments. 

For Inception v3 models, we pre-train from random initialization for 2,000,000 steps using Stochastic Gradient Descent (SGD) with Nesterov momentum. Each mini-batch contained 1,024 examples. The same weight regularization and learning rate parameters were used for all pre-trained models and were selected based on a separate hold-out dataset. We used a learning rate schedule that first starts with a linear ramp up for 20,000 steps, followed by cosine decay. 

AmoebaNet-B models followed a similar setup with pre-training from random initialization for 250,000 steps using SGD and Nesterov momentum. We used larger mini-batches of 2,048 examples to speed up training. The same weight regularization and learning rate parameters were used for all models, and matched the parameters that Real \etal~\cite{Real2018} used for ImageNet training. We chose to use AmoebaNet-B with settings (N=18, F=512), resulting in over 550 million parameters when trained on ImageNet, so as to evaluate our methods on a large model.

During fine-tuning, we used a randomly initialized classification layer in place of the pre-trained classification layer. Models were trained for 20,000 steps using SGD with momentum. Each mini-batch contained 256 examples. The weight regularization and learning rate parameters were determined using a hold-out validation set. We used a similar learning rate schedule with a linear ramp for 2,000 steps, followed by cosine decay.

For domain adaptive transfer learning, we found that adding a smooth prior when computing $P_t(y)$ helped performance with ImageNet as a source pre-training data. Hence, we used a temperature\footnote{The logits are divided by the temperature before computing the softmax.} of $2.0$ when computing the softmax predictions for the computation of the importance weights.

\subsection{Pre-training setup}
\label{section:pre_training_setup}

While it is possible to directly perform pre-training with importance weights, we found it challenging as the importance weights varied significantly. When pre-training on a large dataset, this means that it is possible to have batches of data that are skewed in their weights with many examples weighted lightly. This is also computationally inefficient as the examples with very small weights contribute little to the gradients during training.

Hence, we created pre-training datasets by sampling examples from the source dataset using the importance weights. We start by choosing a desired pre-training dataset size, often large. We then sample examples from the source dataset at a rate proportional to the importance weights, repeating examples as needed. We report results that construct a pre-training dataset of 80 million examples for \largedatasetname, and 2 million examples for ImageNet.  We used the same sampled pre-training dataset with both the Inception v3 and AmoebaNet-B experiments.

\subsection{Transfer learning results}
\label{section:data_matching_results}

\begin{figure*}[!htb]
  \centering
  \includegraphics[width=0.95\textwidth]{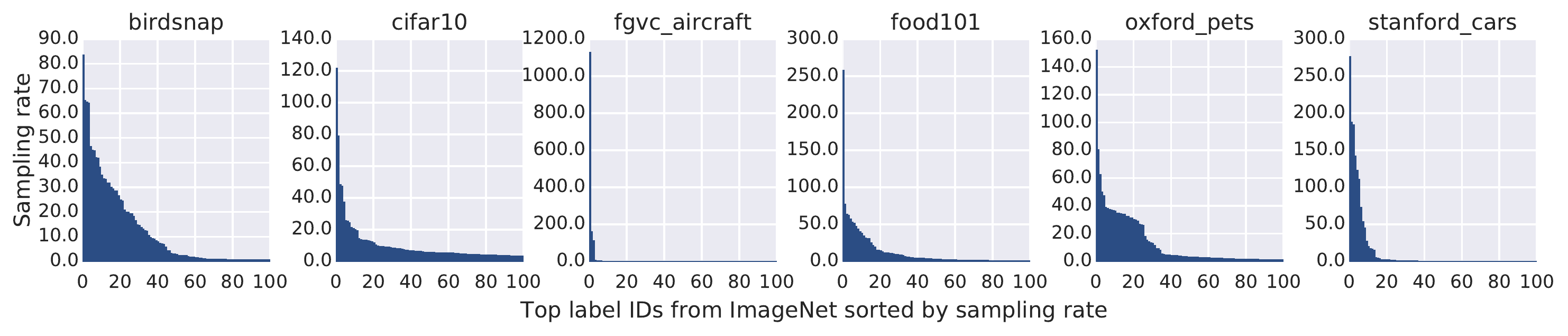}
  
  \caption{Distribution of importance weights for each target dataset when using ImageNet as a source pre-training dataset. The horizontal axis represents the top 100 ImageNet labels sorted by importance weight for each dataset; each dataset has a different order. The distributions vary widely between target datasets. FGVC Aircraft selects only a few labels that turn out to be coarse grained, whereas Oxford Pets selects a wider variety of fine-grained labels. This reflects the inherent bias in the ImageNet dataset.}
  \label{fig:imagenet_subset_distribution}
\end{figure*}

\begin{figure*}[htb]
  \centering
  \includegraphics[width=1.0\textwidth]{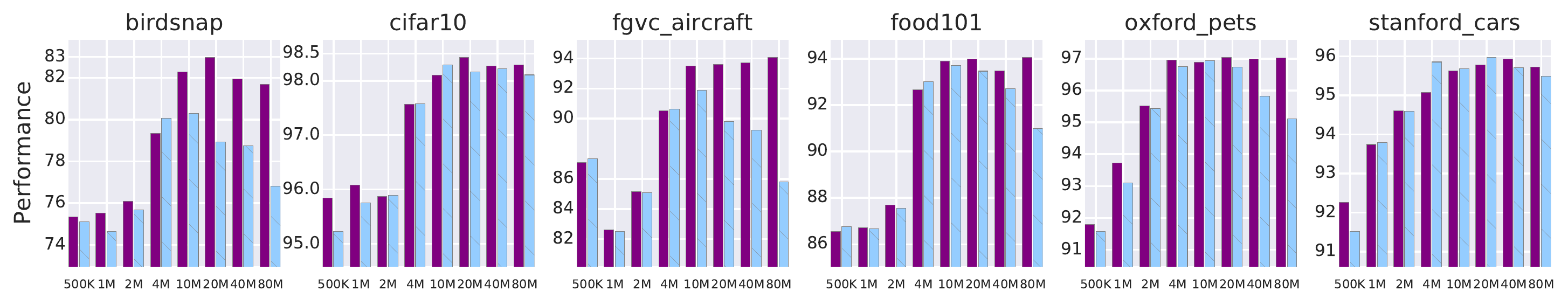}
  \includegraphics[width=1.0\textwidth]{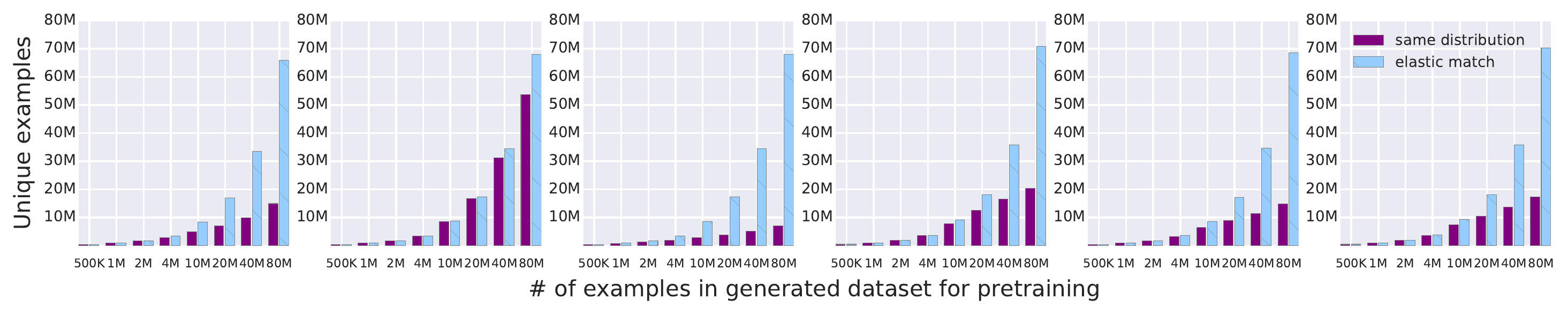}
  
  \caption{Performance (top) and unique examples (bottom) of the same distribution matcher and elastic distribution matcher at different sampled dataset sizes. We see  that when dataset size increases, the performance of same distribution matcher increases and then saturates, while that of elastic distribution matcher drops after a peak. Notice that the elastic distribution matcher also has significantly more unique examples than same distribution matcher as the dataset size increases.}
  \label{fig:distribution_match_by_size}
\end{figure*}

\paragraph{Domain adaptive transfer learning is better.}

When the source pre-training domain matches the target dataset, such as in \largedatasetname-Bird to Birdsnap or \largedatasetname-Cars to Stanford Cars, transfer learning is most effective (Table \ref{table:transfer_results}). However, when the domains are mismatched, we observe negative transfer: \largedatasetname-Cars fine-tuned on Birdsnap performs poorly. Strikingly, this extends to categories which are intuitively close: aircrafts and cars. The features learned to discriminate between types of cars does not extend to aircrafts, and vice-versa. 

\paragraph{More data is not necessarily better.}

Remarkably, more data during pre-training can hurt transfer learning performance. In all cases, the model pre-trained on the entire \largedatasetname~dataset did worse than models trained on more specific subsets.  These results are surprising as common wisdom suggests that more pre-training data should improve transfer learning performance if generic features are learned. Instead, we find that it is important to determine how relevant additional data is.

The ImageNet results with Domain Adaptive Transfer further emphasize this point. For ImageNet with Adaptive Transfer, each pre-training dataset only has around 450k unique examples. While this is less than half of the full ImageNet dataset of 1.2 million examples, the transfer learning results are slightly better than using the full ImageNet dataset for many of the target datasets.

\paragraph{Domain adaptive transfer is effective.}

When pre-training with \largedatasetname~and ImageNet, we find that the domain adaptive transfer models are better or competitive with manually selected labels from the hierarchy. For datasets that are composed of multiple categories such as CIFAR-10 which includes animals and vehicles, we find further improved results since the constructed dataset includes multiple different categories. 

In Figure \ref{fig:imagenet_subset_distribution}, we observe that the distributions are much more concentrated with FGVC Aircraft and Stanford Cars: this arises from the fact that ImageNet has only coarse-grained labels for aircraft and cars. In effect, ImageNet captures less of the discriminative factors of variation that is captured in either FGVC Aircraft and Stanford Cars. Hence, we observe that transfer learning only improves the results slightly.

\begin{table*}[!htb]
\centering
\begin{tabular}{l | c c c c c c} 
 \toprule
 & \multicolumn{6}{c}{Target Fine-tuned Dataset} \\
 Pre-training Method & Birdsnap & Oxford-IIIT & Stanford  & FGVC      & Food-101 & CIFAR-10 \\ 
                      &       & Pets     & Cars      & Aircraft         &          & \\ 
 \midrule
 Entire \largedatasetname~Dataset &  80.3  &  94.5    & 95.3   &  90.5         &  92.0       &  98.6   \\ 
 \largedatasetname~- Bird       &  \textbf{85.5} & 90.4 & 92.0 &  86.9         &  90.7       &  97.8   \\ 
 \largedatasetname~- Animal     &  84.1    & 96.4 &  93.2      &  90.0         &  92.3       &  98.8   \\ 
 \largedatasetname~- Car        &  79.0    &  88.9 & \textbf{96.2} &  92.2     &  90.1       &  96.7   \\ 
 \largedatasetname~- Aircraft   &  78.0    &  87.7 &  93.3     &  92.5         &  89.8       &  97.2   \\ 
 \largedatasetname~- Vehicle    &  78.8    &  88.6 &  96.0     &  93.0         &  90.4       &  97.2   \\ 
 \largedatasetname~- Transport  &  79.2    &  89.1 &  95.9     &  93.1         &  90.4       &  97.3   \\ 
 \largedatasetname~- Food       &  79.7    &  89.2 &  92.6     &  88.7        &   95.1       &  97.5   \\ 
 \largedatasetname~- Adaptive Transfer    &  85.1   &  \textbf{96.8}   &  95.8   &  92.8   &  \textbf{95.3}   &  98.6  \\
 \hline \\ [-0.8em]
 ImageNet - Entire Dataset &  80.8 &  94.5 &  94.2   & 90.7 & 91.7       & 98.0   \\
 ImageNet - Adaptive Transfer &  80.7 &  95.1 &  93.5   &  89.2        &  91.5       &  98.0   \\
 \hline \\ [-0.8em]
 Best Published Results &  83.9 \cite{Krause2016}  & 95.9 \cite{Huang2018}  &  94.6 \cite{Huang2018}   &  \textbf{94.5} \cite{Krause2016}   &  93.0 \cite{Huang2018}       &  \textbf{99.0} \cite{Huang2018} \\ 
\bottomrule
\end{tabular} 
\vspace{1.0em}
\caption{
Transfer learning results with AmoebaNet-B. Each row corresponds ot a pre-training method. Adaptive transfer refers to our proposed method. Results reported are top-1 accuracy for all datasets except Oxford-IIIT Pets, where we report mean accuracy per class. All our results are averaged over 5 fine-tuning runs. Huang~\etal\cite{Huang2018} also use the AmoebaNet-B model, but with a larger input image size at $480\times480$, while we used $331\times331$ image inputs instead.
}
\label{table:transfer_results_amoebanet}
\end{table*}

\subsection{Comparing pre-training sampling mechanisms}
\label{section:compare_dist_matcher}

In section \ref{section:pre_training_setup}, we described a method to construct pre-training datasets from sampling the source dataset. This process also allows us to study the effect of different distributions. Rather than sampling with replacement, as we did earlier, we could also sample \textit{without} replacement when constructing the pre-training dataset. When sampling without replacement, we deviate from the importance weights assigned, but gain more unique examples to train on. We compare these two methods of sampling: (a) sampling with replacement - `same distribution matcher', and (b) sampling without replacement - `elastic distribution matcher'.

When sampling with replacement, we sample examples at a rate proportional to the importance weight computed before, repeating examples as needed. When sampling without replacement, we avoid selecting each example more than once. In order to keep the distribution as similar to the desired one, we consider a sequential approach: we start with the class with the highest importance weight and select all the samples available. Next, we recursively consider sampling a dataset of the remaining desired examples (original desired dataset size minus the number of examples just selected) from the rest of the classes.

By comparing model performances between these two sampling methods, we are able to study how the transfer learning performance varies when (a) there are more unique examples in pre-training, (b) when the distribution in pre-training deviates from the importance weights.

\subsubsection{Comparing sampling methods}

We find that the performance of the same distribution matcher increases, and then saturates. Conversely, the elastic distribution matcher performance first increases then decreases. Note that at the low end of the dataset sizes, both methods will generate similar datasets. Thus, the later decrease in performance from the elastic distribution matcher comes from diverging from the original desired distribution. This indicates that using the importance weights during pre-training is more important than having more unique examples to train on.

\subsection{Results on large models}

We studied our method on large models to understand if large models are better able to generalize, because the increased capacity enables them to capture more factors of variation. We conducted the same experiments on AmoebaNet-B, with over 550 million parameters.

We found that the general findings persisted with AmoebaNet-B: (a) using the entire \largedatasetname~dataset was always worse compared to an appropriate subset and (b) our domain adaptive transfer method was better or competitive with the hand selected subsets. 

Furthermore, we find that the large model was also able to narrow the performance gap between the more general subsets and specific subsets: for example, the performance on Birdsnap between \largedatasetname-Bird and \largedatasetname-Animal is smaller with AmoebaNet-B compared to Inception v3. We also observe better transfer learning between the transportation datasets compared to Inception v3.

Our results are competitive with the best published results (Table \ref{table:transfer_results_amoebanet}). In our experiments, the performance of the AmoebaNet-B was also better in all cases than Inception v3, except for the FGVC Aircraft dataset. This is consistent with Kornblith \etal~\cite{Kornblith2018} who also found that Inception v3 did slightly better than NasNet-A \cite{Zoph2017}. 

\subsection{Comparisons to data selection methods}

Cui \etal~\cite{Cui2018} and Ge \& Yu \cite{Ge17} recently proposed methods for improving transfer learning by selecting relevant data from the source pre-training dataset. We compared our methods to their reported results on the Oxford Flowers 102 \cite{Nilsback2008} dataset, following the same experimental setup in Section \ref{section:pre_training_setup}, using the Inception v3 model.

\begin{table}[!htb]
\centering
\begin{tabular}{l | c} 
 \toprule
 Pre-training Method & Test Accuracy \\
 \hline \\ [-0.8em]
 Entire ImageNet Dataset & 97.1 \\
 ImageNet - Adaptive Transfer & \textbf{97.7} \\
 \hline \\ [-0.8em]
 ImageNet \& iNaturalist - \\ EMD Subset \cite{Cui2018} & \textbf{97.7} \\
 ImageNet - Selective Joint FT \cite{Ge17}  & 97.0 \\
\bottomrule
\end{tabular} 
\vspace{1.0em}
\caption{Transfer learning performance on Oxford Flowers 102 with Inception v3. We report the mean per class accuracy, averaged over 5 runs. The first two rows are our results, showing that adaptive transfer improves over training over the entire ImageNet dataset. Our method performs as well as Cui \etal~\cite{Cui2018} despite using less pre-training data.}
\label{table:compare_data_selection_method}
\end{table}

Our results (Table \ref{table:compare_data_selection_method}) show that our adaptive transfer method can effectively select appropriate examples to pre-train on. We observe an absolute gain of about 0.6\% in accuracy, achieving results comparable to Cui \etal~\cite{Cui2018} despite using less pre-training data.

\subsection{Understanding the importance of the pre-training distribution}

\begin{table}[htb]
\centering
\begin{tabular}{c c} 
 \toprule
 Weighting Scheme &  Transfer Performance \\
 \midrule
 Adaptive Transfer Weights  & 97.1  \\ 
 Uniform Weights  & 95.3 \\
 Reversed Weights  & 84.9 \\ 
 \bottomrule
\end{tabular}
\vspace{1.0em}
\caption{Transfer performance on Oxford-IIIT Pets from \largedatasetname~subsets of the same size (80M), using the top 4k selected labels, but with different weighting schemes: \textit{Adaptive Transfer Weights} are the weights from our method, \textit{Uniform} is a uniform distribution on all selected labels, and \textit{Reverse} is a distribution obtained by swapping the weights between the highest and the lowest weighted labels from the \textit{Adaptive Transfer} distribution.}
\label{table:transfer_perf_different_distribution}
\end{table}

In contrast to Cui \etal~\cite{Cui2018}, our method not only selects the relevant categories from the source pre-training set, but also weights them. In order to understand the importance of the weighting scheme, we performed experiments to vary the weights assigned over a subset of selected categories. 

We created \largedatasetname~subsets over the top 4,000 matched labels on Oxford-IIIT Pets.  We used three different weighting schemes in our comparison: (a) importance weights according to our adaptive transfer method, (b) uniform weights across all the labels, and (c) adaptive transfer weights reversed, where we assigned the highest weight to the label with originally the lowest weight. 

Our results (Table \ref{table:transfer_perf_different_distribution}) show that uniform weights are inferior to the adaptive transfer weights, indicating that is it important to emphasize some labels more than others. The reversed weights perform the worst.

\section{Discussion}

Transfer learning appears most effective when the pre-trained model captures the discriminative factors of variation present in the target dataset. This is reflected in the significant overlap in the classes between ImageNet and other datasets such as Caltech101, CIFAR-10, etc. where transfer learning with ImageNet is successful. Our domain adaptive transfer method is also able to identify the relevant examples in the source pre-training dataset that capture these discriminative factors.

Conversely, the cases where transfer learning is less effective are when it fails to capture the discriminative factors. In the case of the ``FGVC Aircraft" dataset \cite{Maji2013}, the task is to discriminate between 100 classes over manufacturer and models of aircraft (e.g., Boeing 737-700). However, ImageNet only has coarse grained labels for aircraft (e.g., airliner, airship). In this case, ImageNet models tend to learn to ``group" different makes of aircraft together rather than differentiate them. It turns out that the \largedatasetname~dataset has fine-grained labels for aircraft and is thus able to demonstrate better transfer learning efficacy.

Our results using AmoebaNet-B show that even large models transfer better when pre-trained on a subset of classes, suggesting that they make capacity trade-offs between the fine-grained classes when training on the entire dataset. This finding posits new research directions for developing large models that do not make such a trade-off. 

We have seen an increase in dataset sizes since ImageNet; for example, the YFCC100M dataset \cite{Thomee2016} has 100M examples. We have also seen developments of more efficient methods to train deep neural networks. Recent benchmarks \cite{Coleman2018} demonstrate that it is possible to train a ResNet-50 model in half an hour, under fifty US dollars. This combination of data and compute will enable more opportunities to employ better methods for transfer learning.

\newpage
\subsection*{Acknowledgments}

We wish to thank Yanping Huang, Jon Shlens, Sergey Ioffe, Tom Duerig, Tomas Pfister, Yin Cui, Vishy Tirumalashetty, and Chen Sun for helpful feedback and discussions.

{\small

\bibliographystyle{ieee}
}

\end{document}

%% file: math_commands.tex
\usepackage{amsmath,amsfonts,bm}

\def\eqref#1{equation~\ref{#1}}

\def\1{\bm{1}}

\DeclareMathAlphabet{\mathsfit}{\encodingdefault}{\sfdefault}{m}{sl}
\SetMathAlphabet{\mathsfit}{bold}{\encodingdefault}{\sfdefault}{bx}{n}

